\documentclass[letterpaper, 10 pt, journal, twoside]{IEEEtran} 

\IEEEoverridecommandlockouts 
\usepackage{color}
\newcommand{\ie}{i.e.\ }

\newcommand{\Reffig}[1]{Figure~\ref{#1}}
\newcommand{\Refsec}[1]{Section~\ref{#1}}
\newcommand{\Refalg}[1]{Algorithm~\ref{#1}}

\newcommand{\Reftab}[1]{Table~\ref{#1}}

\usepackage{graphics} 
\usepackage{epsfig} 
\usepackage{url} 
\usepackage{algorithm}
\usepackage{algpseudocode}
\usepackage{amsmath}
\usepackage{amssymb}
\usepackage{subfigure}
\usepackage{graphicx}
\usepackage{subfigure}
\usepackage{booktabs}
\usepackage{tabularx}
\usepackage{threeparttable}
\usepackage{multirow}
\usepackage{multicol}
\usepackage{makecell}
\usepackage{bm}
\usepackage{makecell}
\usepackage{pifont}
\usepackage{cite}
\usepackage{soul}
\def\IEEEkeywordsname{Index Terms}

\title{N$^{3}$-Mapping: Normal Guided Neural Non-Projective Signed Distance Fields for Large-scale 3D Mapping}
\author{Shuangfu Song$^{1}$, Junqiao Zhao$^{*, 2}$, Kai Huang$^{1}$, Jiaye Lin$^{2}$, Chen Ye$^{2}$, Tiantian Feng$^{1}$
    \thanks{Manuscript received: December, 31, 2023; Revised: March, 28, 2024; Accepted: April, 27, 2024.}
    \thanks{This paper was recommended for publication by Editor Sven Behnke upon evaluation of the Associate Editor and Reviewers' comments.
    This work is supported by the National Key Research and Development Program of China (No. 2021YFB2501104). \emph{(Corresponding Author: Junqiao Zhao.)}} 
    \thanks{$^{1}$Shuangfu Song, Kai Huang and Tiantian Feng are with the School of Surveying and Geo-Informatics, Tongji University, Shanghai 200092, China,
        {E-mail: \tt\footnotesize \{songshuangfu, huangkai, fengtiantian\}@tongji.edu.cn.}}
    \thanks{$^{2}$Junqiao Zhao, Jiaye Lin and Chen Ye are with the Department of Computer Science and Technology, School of Electronics and Information Engineering, Tongji University, Shanghai 201804, China,
        {E-mail: \tt\footnotesize \{zhaojunqiao, jiayelin, yechen\}@tongji.edu.cn.}}
    \thanks{Digital Object Identifier (DOI): see top of this page.}
}

\begin{document}
\maketitle
\begin{abstract}
Accurate and dense mapping in large-scale environments is essential for various robot applications.
Recently, implicit neural signed distance fields (SDFs) have shown promising advances in this task.
However, most existing approaches employ projective distances from range data as SDF supervision, introducing approximation errors and thus degrading the mapping quality.
To address this problem, we introduce N$^{3}$-Mapping, an implicit neural mapping system featuring normal-guided neural non-projective signed distance fields.
Specifically, we directly sample points along the surface normal, instead of the ray, to obtain more accurate non-projective distance values from range data.
Then these distance values are used as supervision to train the implicit map.
For large-scale mapping, we apply a voxel-oriented sliding window mechanism to alleviate the forgetting issue with a bounded memory footprint.
Besides, considering the uneven distribution of measured point clouds, a hierarchical sampling strategy is designed to improve training efficiency.
Experiments demonstrate that our method effectively mitigates SDF approximation errors and achieves state-of-the-art mapping quality compared to existing approaches.
The code will be released at \url{https://github.com/tiev-tongji/N3-Mapping}.
\end{abstract}


\section{INTRODUCTION}
\IEEEPARstart{B}{uilding} accurate and dense maps within large-scale environments is crucial for various robot applications, such as autonomous driving.
The signed distance field (SDF) has been an important map representation employed to accomplish this task, due to its advantages in characterizing geometric information and compatibility with downstream tasks \cite{oleynikova2016sdf}.
Unfortunately, traditional SDF mapping methods \cite{oleynikova2017Voxblox,vizzo2022VDBFusion, izadi2011Kinect} have faced two longstanding challenges.
The first is the trade-off between mapping accuracy and memory consumption.
The second is the approximation error introduced by using projective distance, \ie the distance along range sensor rays to the measured surface, to estimate SDF values.
This can easily lead to an overestimation of the actual distance to the nearest surface.

Recently, neural implicit representations \cite{park2019DeepSDF,mescheder2019Occupancy,mildenhall2020NeRF} have shown promising potential in modeling 3D scenes.
Neural network-based distance fields exhibit continuity, enabling them to overcome resolution limitations and achieve high-fidelity reconstructions.
However, learning accurate distance fields from raw range sensor data is not trivial, as it is challenging to obtain true distance supervision.
Most current methods \cite{azinovic2022Neural,wang2022GOSurf,yang2022Voxfusion,zhong2023shine,liu2023RIM} still employ projective distance to approximate the ground-truth SDF supervision for training efficiency, while neglecting the associated systematic errors.
Several works \cite{deng2023NeRFLOAM,wiesmann2023LocNDF} leverage surface normals to correct the projective distance and reduce the approximation error.
Nonetheless, the former \cite{deng2023NeRFLOAM} focuses solely on large planar surfaces, limiting its applicability.
The latter \cite{wiesmann2023LocNDF}, on the other hand, lacks training stability as it approximates the normal direction using the unstable gradients of neural networks during training.
Currently, it is vital to develop a practical non-projective SDF mapping system to overcome these limitations.

To this end, we propose N$^{3}$-Mapping, a large-scale mapping approach with normal guided neural non-projective signed distance fields.
We observe that, for any point around the surface, the normal generally provides the direction to the nearest surface point.
Therefore, our method directly samples points and corresponding distance values along the normal direction near the surface.
Such sampled SDF labels tend to be close to the ground truth, leading to improved mapping quality.
For the implicit map, our method utilizes octree-based hierarchical sparse voxels to store optimizable feature vectors and a shallow MLP to decode queried local features into SDF values.

For large-scale mapping, we employ a voxel-oriented training strategy to alleviate catastrophic forgetting by caching historical supervision signals into their corresponding map voxels.
A sliding window mechanism is then applied to promptly drop data outside the window, ensuring a bounded memory footprint.
To further improve efficiency, we propose a hierarchical sampling strategy to avoid redundant training in densely observed regions and insufficient training in sparsely observed regions.

Our method achieves efficient, high-quality and incremental mapping in large-scale environments. To summarize, our contributions are as follows:
\begin{itemize}
    \item[1] A simple yet effective neural non-projective SDF training method guided by surface normals, enables accurate and complete 3D dense mapping.
    \item[2] A voxel-oriented training strategy combined with a sliding window mechanism and hierarchical sampling, mitigates the forgetting issue and enhances training efficiency.
    \item[3] Extensive experiments demonstrate that our method outperforms existing approaches in terms of mapping accuracy and completeness.
\end{itemize}
\section{RELATED WORK} \label{sec2}
Various scene representations have been explored for 3D dense mapping, including occupancy grids \cite{elfes1989occupancy}, meshes \cite{lorensen1997Marching} and surfels \cite{pfister2000surfels}.
In addition to these, SDFs have gained significant popularity in many robotic tasks such as planning \cite{zucker2013CHOMP,oleynikova2017Voxblox}, localization \cite{huang2019Metric, wiesmann2023LocNDF}, and 3D mapping \cite{izadi2011Kinect,vizzo2022VDBFusion}.
As a milestone of using truncated signed distance function (TSDF) as map representation, KinectFusion \cite{izadi2011Kinect} uses depth images as input and introduces the volumetric integration algorithm to enable real-time mapping.
Subsequent studies follow the manner of TSDF integration and work on enhancing efficiency \cite{oleynikova2017Voxblox, vizzo2022VDBFusion}, accuracy \cite{pan2022Voxfield}, and the capability to handle dynamic \cite{newcombe2015dynamic} and large-scale environments \cite{whelan2015large}.
However, constrained by their explicit representations, these methods require substantial memory resources to achieve high resolutions.

Recently, research on implicit neural representations has made significant progress.
Seminal works \cite{park2019DeepSDF, mescheder2019Occupancy, mildenhall2020NeRF} use implicit neural networks to represent 3D objects and scenes, overcoming the limitations of traditional explicit representations. 
Inspired by them, \cite{sucar2021imap,ortiz2022isdf,azinovic2022Neural} achieve incremental neural mapping by continuously training an MLP to represent the environment with sequentially input data. 
Considering the limited capacity of a single MLP, NICE-SLAM \cite{zhu2022nice} combines dense grids to store optimizable local features and employs shallow MLPs to decode hidden information.
This combination allows for more accurate and faster reconstruction but dense grids are not memory-efficient.
Subsequent works tackle this issue by employing various data structures such as sparse octree \cite{zhong2023shine,yang2022Voxfusion}, hash encoding \cite{muller2022Instant}, and neural points \cite{sandstrom2023PointSLAM}.
In our approach, we adopt a sparse octree to store multi-layer local features, as done in SHINE-Mapping \cite{zhong2023shine}.

For training SDF with range data, recent studies \cite{azinovic2022Neural,wang2022GOSurf,yang2022Voxfusion,zhong2023shine,liu2023RIM} commonly use projective distance, \ie the signed distance from a sampled point along the ray to the endpoint, as the supervision signal.
This SDF approximation can lead to faster convergence while it also introduces errors, especially when the incidence angle of the ray is small.
To obtain more accurate non-projective SDF, Voxfield \cite{pan2022Voxfield} performs normals integration for each voxel and uses the fused normal to correct the distance value.
Similarly, NeRF-LOAM \cite{deng2023NeRFLOAM} applies projective distance correction with normals specifically for ground points, and LocNDF \cite{wiesmann2023LocNDF} uses the gradient of the distance field to rectify SDF labels.
Nevertheless, NeRF-LOAM only considers the ground plane, and LocNDF faces training instability due to circular dependency.
Additionally, \cite{ortiz2022isdf,shi2023Accurate} compute the distance between sampled points and the nearest measured surface points as the SDF label.
However, such methods are sensitive to noise and rely on pre-built dense point clouds \cite{shi2023Accurate}.
In our method, accurate SDF labels can be easily obtained by directly sampling points along the normal direction.

For incremental mapping with implicit representations, it has been shown that both neural network-based methods and feature grid-based methods face the challenge of catastrophic forgetting \cite{zhong2023shine}.
Most of these approaches \cite{sucar2021imap,zhu2022nice,ortiz2022isdf,yang2022Voxfusion,azinovic2022Neural} mitigate this issue by replaying historical keyframes.
However, such replay-based methods are memory inefficient because they require storing historical keyframes, making them challenging to handle large-scale environments.
To solve this problem, SHINE-Mapping \cite{zhong2023shine} designs an update regularization strategy, albeit with a decrease in mapping performance.
RIM \cite{liu2023RIM} prioritizes robot-centric local implicit maps to achieve more efficient training but decouples the local map and global map.
In contrast, our approach seamlessly integrates a voxel-oriented sliding window into the global implicit map.
Supervisions from different viewpoints are accumulated in their respective voxels for batch training, resulting in smooth and consistent mapping results.
Additionally, existing methods typically rely on naive random sampling to select samples for training, leading to redundant computations in densely observed regions while somewhat neglecting sparsely observed regions \cite{jiang2023H2}.
Our hierarchical sampling strategy ensures that all regions receive appropriate training attention.

\section{N$^{3}$-Mapping} \label{method}

\begin{figure}
    \centering
    \includegraphics[width=0.95\linewidth]{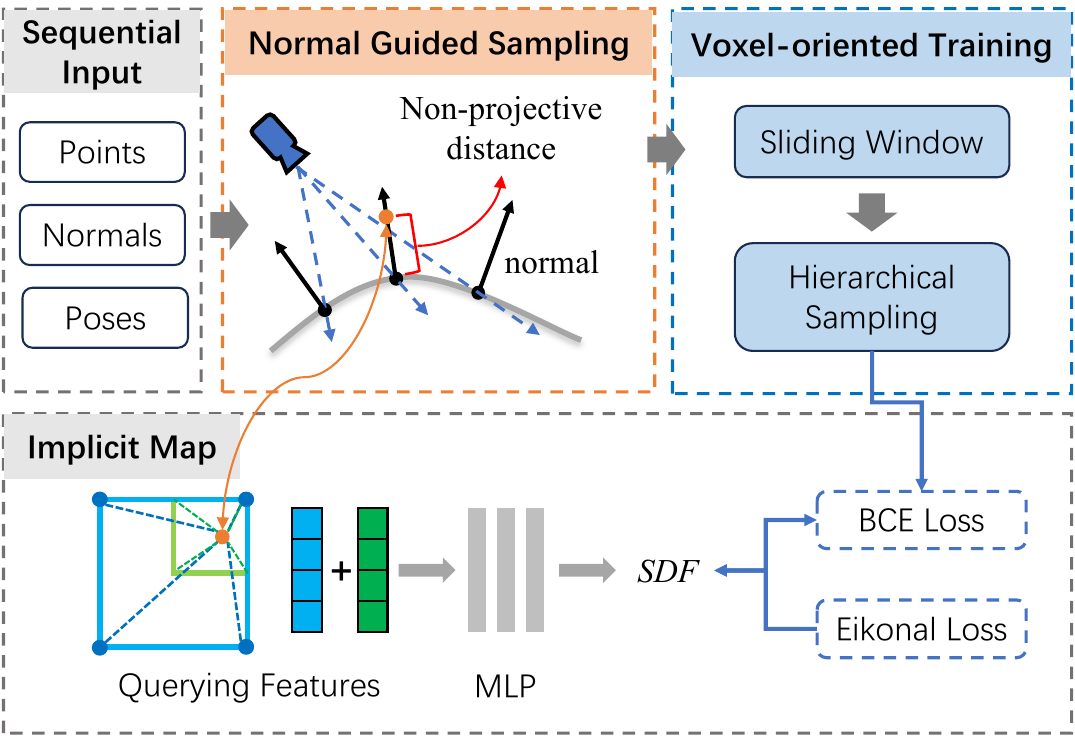}
    \caption{Overview of our approach. With a sequence of range data and corresponding poses, our approach samples non-projective distance values to obtain accurate SDF labels along the normal direction.
    During training, these labels are used to supervise the learning of our implicit map through the voxel-oriented training strategy.}
    \label{fig_overview}
    \vspace{-0.4cm}
\end{figure}
In this paper, we present N$^3$-Mapping, a framework designed for large-scale and high-quality 3D mapping. as outlined in \Reffig{fig_overview}.
Given sequentially input points and normals from LiDAR or RGB-D cameras with associated poses, we obtain accurate non-projective SDF labels, along with corresponding sampled points, through normal-guided sampling (\Refsec{sec_np}).
These training pairs are then maintained in our voxel-oriented sliding window (\Refsec{sec_sw}).
For subsequent training, by employing a hierarchical sampling strategy (\Refsec{sec_hier}), we select a batch of training pairs at each iteration to optimize our octree-based implicit map (\Refsec{sec_map}).
Finally, we visualize and evaluate the mapping results as 3D meshes generated through the marching cubes algorithm \cite{lorensen1997Marching}.

\subsection{Preliminaries}
\subsubsection{Signed Distance Fields} \label{sec_sdf}
A signed distance field represents the 3D scene by assigning a signed distance to each point in space relative to the closest surface.
The sign of the distance indicates whether a point is inside or outside the surface.
The signed distance function \(f\) can be defined as: \(f(\mathbf{x}) = s\), which maps a 3D coordinate \(\mathbf{x}\) to the signed distance value \(s\).

Notably, the gradient of SDF $\nabla_\mathbf{x} f(\mathbf{x})$ points towards the closest surface and on the surface the gradient is equal to the surface normal: $\nabla_\mathbf{x} f(\mathbf{x}) = \mathbf{n}$ (or \(-\mathbf{n}\) depending on the normal's orientation).
This implies that normal priors can provide informative guidance for neural SDF learning.

\subsubsection{Implicit Neural Map Representation} \label{sec_map}
In this paper, we utilize octree-based hierarchical sparse voxels to store learnable features at each node vertex and a shallow MLP to decode hidden features into SDF values.
Following SHINE-Mapping \cite{zhong2023shine}, we build a hash table with Morton codes as keys to query features efficiently and facilitate map scalability.
For an arbitrary point $\mathbf{x}$ within the map, its corresponding local feature vector at a given level $l$ can be obtained through trilinear interpolation function $TriLerp(\cdot)$ with its eight neighboring feature vectors $\{\mathbf{v}_k^l\}$.
Then different levels of features are aggregated by summation and then fed into a shallow MLP $f_\theta$ with globally shared weights to decode the SDF value $s$:
$$
s = {f_\theta}(\sum_{l = 1}^{L} TriLerp ( \mathbf{x}, \{\mathbf{v}_i^l\})), \quad i \in \{1, \ldots, 8\}.
$$
We denote this function as $s = f_\theta(\mathbf{x})$ for brevity in the following sections.
Since the entire process is differentiable, we can optimize the feature vectors and the parameters of the MLP jointly using the generated signed distance value as supervision.

\subsection{Non-projective SDF Learning} \label{sec_np}
\begin{figure} 
    \centering
    \includegraphics[width=0.95\linewidth]{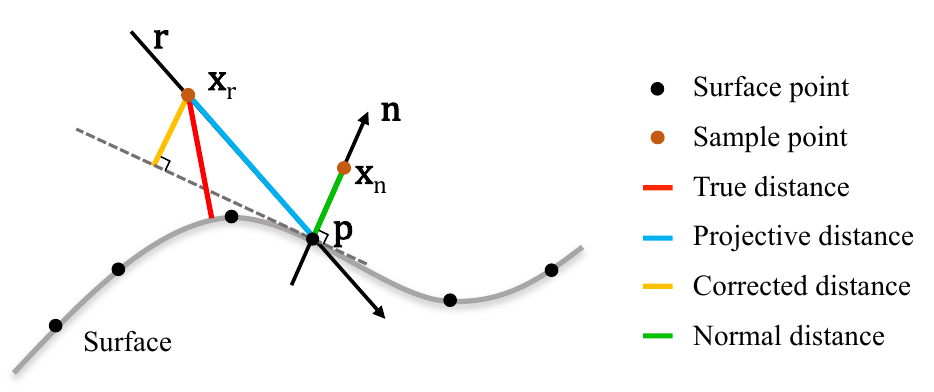}
    \caption{Different methods for sampling signed distance values.
        Both projective distance along the ray and corrected distance with normals lead to large errors on curved surfaces.
        Our normal-guided sampling method can produce more accurate distance values.}
    \label{fig_np}
    \vspace{-0.5cm}
\end{figure}

\subsubsection{Normal Guided Non-projective Distance} \label{sec_normal}
Ideally, we aim to use true signed distance values to supervise the learning of our implicit map.
As discussed in \Refsec{sec2}, existing methods often resort to projective distance as SDF approximation, which is obtained by sampling points along the ray $\mathbf{r}$ and calculating the distance $s$ from the sampled point \(\mathbf{x}_r\) to the endpoint \(\mathbf{p}\):
\[
s = sign(\mathbf{r} \cdot (\mathbf{p - x}_r)) ||\mathbf{p - x}_r||.
\]
This distance can be further projected along the surface normal $\mathbf{n}$ to reduce the approximation error \cite{deng2023NeRFLOAM,wiesmann2023LocNDF}:
\[
s = sign(\mathbf{r} \cdot (\mathbf{p - x}_r)) (\mathbf{p - x}_r) \cdot \mathbf{n}.
\]
However, 
this solution struggles to deal with curved or irregular surfaces, which are common in real-world scenarios.
As illustrated in \Reffig{fig_np}, both the projective distance (blue line) and corrected distance (yellow line) still remain a substantial gap to the true distance (red line).

As discussed in \Refsec{sec_sdf}, along the negative normal direction, the sampled points direct toward the closest surface, aligning with the gradient direction of the distance field.
Therefore, we propose to directly sample points along the normal direction $\mathbf{n}$ and use the signed distance from the sampled point $\mathbf{x}_n$ to the surface point $\mathbf{p}$ as the supervision:
\[
s = sign(\mathbf{n} \cdot (\mathbf{p - x}_n)) ||\mathbf{p - x}_n||.
\]
Such sampled SDF values are quite close to the ground truth.
However, as the sampling distance increases, this approximation becomes less reliable because some sampled points may be closer to other surfaces.
Hence, we set a truncation interval \([-tr, tr]\) around the surface and only perform normal-guided sampling inside this region according to the Gaussian distribution \(\mathcal{N}(0, \sigma)\), where $\sigma$ is a hyperparameter that indicates the magnitude of the measurement noise.
We also sample points in free space along the ray between the sensor and the truncation region.
The SDF labels of these points are set to the truncation value $tr=3\sigma$.
This primarily aims to eliminate potential artifacts caused by dynamic objects, without compromising the quality of surface reconstruction.

\subsubsection{Loss Function} \label{sec_loss}
We apply the binary cross entropy (BCE) loss function like SHINE-Mapping \cite{zhong2023shine} since it inherently enhances the importance of SDF labels that are close to zero.
First, we map the SDF value to the occupancy probability within the range of (0, 1) through the sigmoid function: $S(x) = 1/(1 + e^{-\frac{x}{\beta}})$, where $\beta$ can control the reconstruction sharpness.
Given sampled point $\mathbf{x}_i$ with SDF label $s_i$, forming a training pair, we use its corresponding occupancy value $o_i = S(s_i)$ as supervision.
Then, the BCE loss can be formulated as follows:
\begin{equation}
    \mathcal{L}_{bce} = -o_i\log(\hat{o}_i) + (1-o_i)\log(1-\hat{o}_i), 
\label{bce_loss}
\end{equation}
where $\hat{o}_i = S(f_\theta(\mathbf{x}_i))$ represents the predicted value of our implicit model after the sigmoid mapping.

Furthermore, if \(\mathbf{x}_i\) is inside the truncation region, we add an eikonal term to ensure valid and continuous SDF values.
This term serves as a regularization to encourage the SDF gradient w.r.t. 3D query point coordinates to have a unit length:
\begin{equation}
    \mathcal{L}_{eik} = (\| \nabla_{\mathbf{x}_i}f_\theta(\mathbf{x}_i)\| - 1)^2.
\end{equation}

In summary, the final loss function is designed as follows:
\begin{equation}
\mathcal{L}_{total} = \mathcal{L}_{bce} + \lambda_e \mathcal{L}_{eik},
\end{equation}
where $\lambda_e$ represents the weight of the eikonal term.
It is worth noting that we do not explicitly supervise the SDF gradient prediction using surface normals as done in \cite{sitzmann2020siren,gropp2020IGR}.
This is because the normal information has already been implicitly encoded in the SDF labels through the preceding normal-guided sampling process.
Consequently, this simplified training process can lead to improved mapping efficiency.

\subsection{Voxel-oriented Training}
\subsubsection{Voxel-oriented Sliding Window} \label{sec_sw}

\begin{figure} 
    \centering
    \vspace{0.2cm}
    \includegraphics[width=0.99\linewidth]{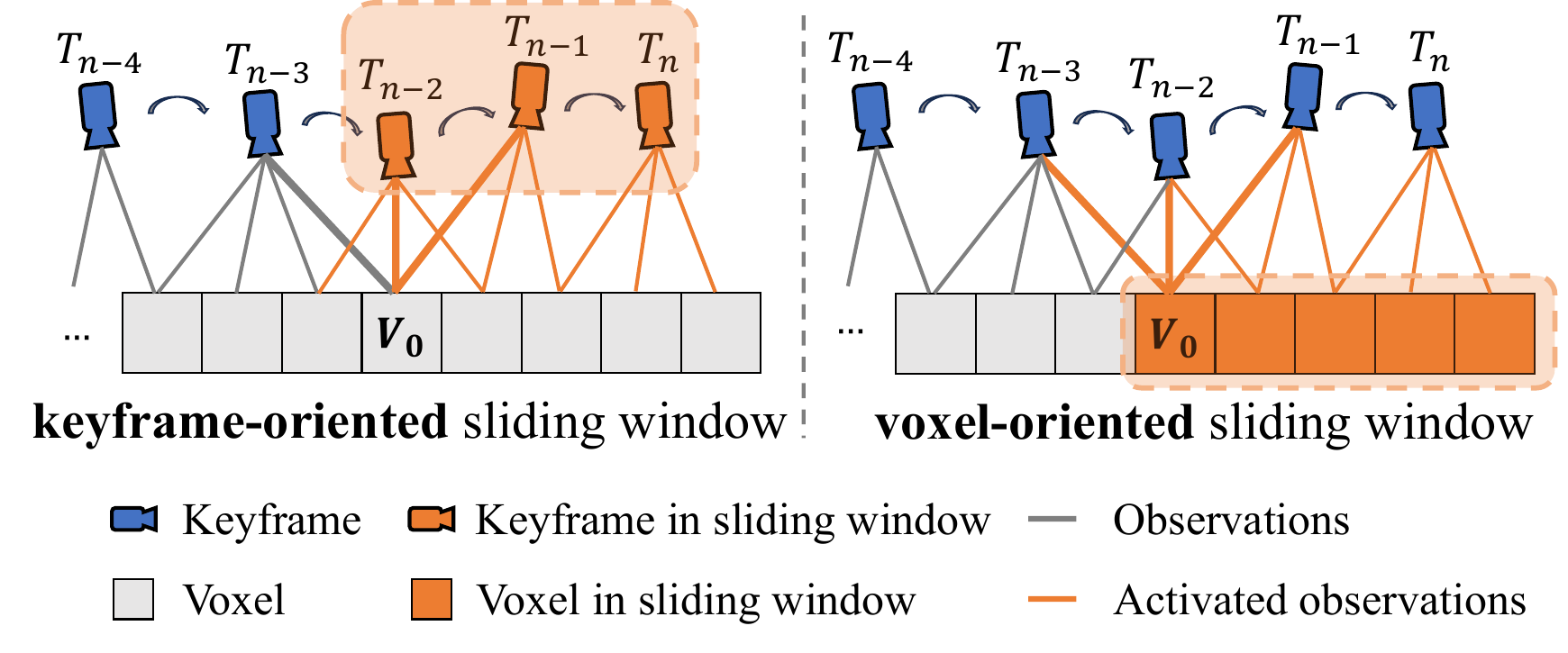}
    \caption{Illustration of the difference between keyframe-oriented and our proposed voxel-oriented sliding window.
    For the voxel $\boldsymbol{V_0}$ at the edge of the local window, our strategy preserves all supervisions from various views, while the keyframe-oriented method retains only partial observations, potentially leading to a forgetting issue.}
    \label{fig_sw}
    \vspace{-0.5cm}
\end{figure}

Current feature grid-based methods \cite{zhu2022nice,yang2022Voxfusion,wang2022GOSurf} often select recently observed keyframes from the global set for efficient local optimization, similar to the sliding window method employed in traditional SLAM systems.
However, as illustrated in \Reffig{fig_sw}, this keyframe-oriented sliding window approach omits observations from frame \(T_{n-3}\) for voxel \(V_0\).
It has been revealed that optimizing local features using such partial observations can lead to non-smooth reconstruction results \cite{zhong2023shine}.

To tackle this issue, we propose a voxel-oriented sliding window strategy, in which each observation is associated with a voxel instead of a keyframe.
Historical data is accumulated in their corresponding voxels by hash querying.
Benefiting from this design, each voxel within the local window can retain complete supervision signals for training.
For simplicity, we define the sliding window as a cube within the octree structure of our implicit map (as detailed in \Refsec{sec_map}).
The local window is centered at the current sensor's origin and its main axes are aligned with the global octree.
Given the sensor origin coordinates $\mathbf{o}$ and the maximum perception range $r$, the 3D bounding box of this cube in voxel units can be derived as follows:
\begin{equation}
\begin{gathered}
    \mathbf{o}^v = \left\lfloor \frac{\mathbf{o}}{v} \right\rfloor, \quad r^v = \left\lfloor \frac{r}{v} \right\rfloor \\
    [\mathbf{b}_u^v, \mathbf{b}_l^v] = [\mathbf{o}^v + r^v, \mathbf{o}^v - r^v],
\end{gathered}
\end{equation}
where $v$ is the leaf voxel size, $\lfloor \cdot \rfloor$ denotes the floor function, $\mathbf{o}^v$ is the voxel coordinate of the sensor origin, $r^v$ is the perception range in voxel units, and $[\mathbf{b}_u^v, \mathbf{b}_l^v]$ represents the upper and lower bounds of the 3D bounding box in the voxel coordinate frame.
This local window allows our approach to focus on mapping newly explored regions with limited memory usage while avoiding the forgetting issue.
Moreover, to ensure global consistency, the MLP decoder will be frozen once convergence is achieved with a certain number of initial frames.

\subsubsection{Hierarchical Sampling} \label{sec_hier}

We note that the point cloud measurements tend to be unevenly distributed on the surface.
If we simply employ random sampling at each training iteration, regions with higher point cloud density are more likely to be trained, while sparsely observed regions lack sufficient optimization to achieve complete convergence.

To address this problem, we adopt a hierarchical sampling strategy.
In our implementation, each voxel maintains a block of training pairs, which includes query points and corresponding SDF labels, stored in an array.
These arrays are organized in the list \( L_P \). 
The association between each voxel and its index in \( L_P \) is maintained through a look-up table \( L \).
As detailed in \Refalg{alg1}, our sampling strategy involves two stochastic stages.
In the first stage, we randomly sample $N_v$ voxels within the sliding window.
This ensures a spatially uniform voxel-wise sampling, independent of the point cloud's distribution.
Then, within each sampled voxel, we further sample $N_p$ training pairs.
If a voxel contains pairs fewer than a threshold $N_t$, we reduce the sampling number until it accumulates more observations.
After collecting all sampled training pairs, we compute loss functions as described in \Refsec{sec_loss}.
In practice, we perform hierarchical sampling in a parallel manner to accelerate this process.

\begin{algorithm}
\caption{Hierarchical Sampling}\label{alg1}
\begin{algorithmic}[1]
\Require List of training pair arrays $L_P$; Voxels IDs (Morton codes) $M$ in the local window; Look-up table $T$; Sampling number $N_v$, $N_p$; Threshold number $N_t$.
\Ensure Sampled subset of training pairs $P_{\text{sub}}$.
\State $P_{\text{sub}} \gets$ \textsc{EmptyArray}
\State $M' \gets$ \textsc{RandSample}($M, N_v$)
\For{$m_i \in M'$}
    \State $\text{idx} \gets T[m_i]$, index of pairs array.
    \State $P_i \gets L_P[\text{idx}]$, get pairs array.
    \If{$\text{len}(P_i) > N_t$}
        \State $P' \gets$ \textsc{RandSample}($P_i, N_p$)
    \Else
        \State $P' \gets$ \textsc{RandSample}($P_i, N_p/3$)
    \EndIf
    \State $P_{\text{sub}} \gets P_{\text{sub}} + P'$
\EndFor
\end{algorithmic}
\end{algorithm}

\section{EXPERIMENTS}

\subsection{Experimental Setup}

\subsubsection{Baselines}
For comparison, we choose two advanced TSDF integration-based methods, Voxblox \cite{oleynikova2017Voxblox} and Voxfield \cite{pan2022Voxfield}, along with two state-of-the-art implicit representation-based methods, SHINE-Mapping \cite{zhong2023shine} and NeRF-LOAM \cite{deng2023NeRFLOAM}, as our baselines.
The odometry module of NeRF-LOAM is omitted since our focus lies in mapping performance.
All methods run incremental mapping using their official open-source code with ground truth poses and the same voxel size.

\subsubsection{Metrics}
To quantitatively evaluate mapping results, we adopt standard reconstruction metrics including Accuracy (Acc., $cm$), Completion (Comp., $cm$), Chamfer-L1 distance (C-L1, $cm$), Completion Ratio (Comp.Ratio, $\%$), and F-score ($\%$).
Considering that the observed regions may not fully cover the ground truth model, the unobserved portions will be culled before evaluation.
The final mesh used for evaluation is generated by marching cubes on the same fixed-size grid to ensure fairness.

\subsubsection{Implementation Details}
Our method takes as input point-wise normals which are estimated using Open3D \cite{zhou2018open3d} with k-nearest neighbors set to 20.
For implicit map representation, feature vectors are stored in the lowest three levels of the octree, each with a length of 8.
The leaf voxel size is set to $0.2m$ and the truncation distance is $0.3m$.
We sample 3 points near the surface and another 3 in free space.
Our shared MLP decoder contains 2 fully-connected layers with ReLU activations, and each layer has 32 hidden units.
For training, we set $Nv = 1024$, $N_p = 8$ and the sharpness parameter $\beta=0.1$.
All experiments are conducted on a desktop PC with a 3.7 GHz Intel i9-10900X CPU and an NVIDIA RTX3090 GPU.

\subsection{Mapping Quality}

\subsubsection{Maicity}
\begin{figure*} 
    \centering
    \includegraphics[width=0.95\linewidth]{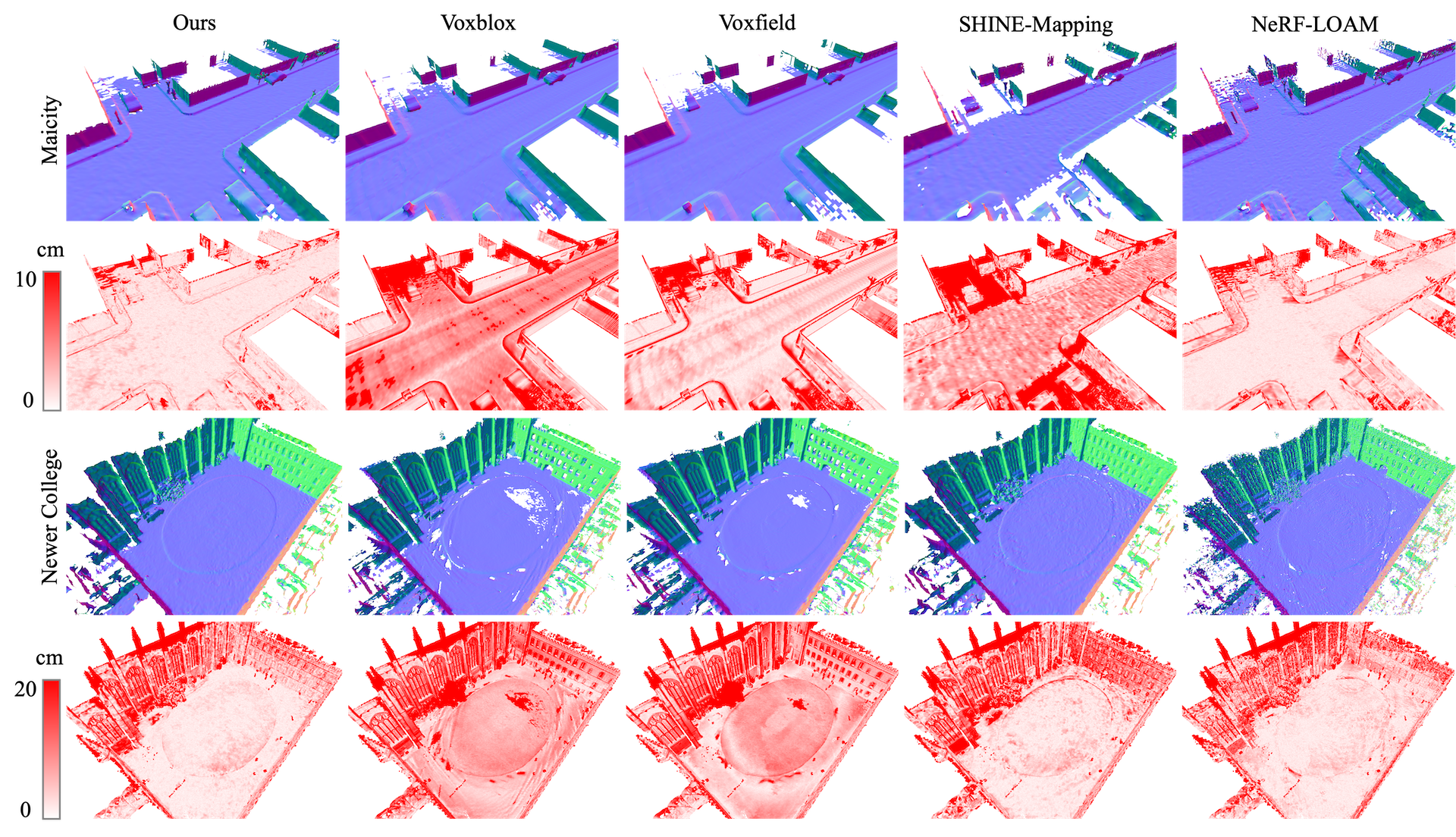}
    \caption{A qualitative comparison of different methods on the \emph{Maicity} and the \emph{Newer College dataset}. The odd rows show the reconstructed mesh colored by surface normals. The even rows present the error map with the ground truth point cloud as a reference where the redder points represent larger errors.}
    \label{fig_qualitative}
    \vspace{-0.3cm}
\end{figure*}

\vspace{0.5cm}
\begin{table}[]
    \caption{Quantitative mapping results of different methods on the \emph{MaiCity dataset}.}
    \vspace{-0.2cm}
    \centering
    \setlength{\tabcolsep}{1.0mm}{
    \begin{tabular}{@{}ccccccc@{}}
    \toprule
    \textbf{Method} & \begin{tabular}[c]{@{}c@{}}\textbf{Acc.} $\downarrow$ \\ {[}$cm${]} \end{tabular} & \begin{tabular}[c]{@{}c@{}}\textbf{Comp.} $\downarrow$ \\ {[}$cm${]} \end{tabular} & \begin{tabular}[c]{@{}c@{}}\textbf{C-L1.} $\downarrow$ \\ {[}$cm${]} \end{tabular} & \begin{tabular}[c]{@{}c@{}}\textbf{Comp.Ratio} $\uparrow$ \\ {[}$10cm,$ $\%${]} \end{tabular} & \begin{tabular}[c]{@{}c@{}}\textbf{F-score} $\uparrow$ \\ {[}$10cm,$ $\%${]} \end{tabular} \\ \midrule
    Voxblox         & 4.76                                                             & 18.92                                                             & 11.84                                                            & 69.65                                                                  & 80.24                                                               \\
    Voxfield        & 3.64                                                             & 11.84                                                             & 7.74                                                             & 82.76                                                                  & 88.78                                                               \\
    SHINE           & 3.82                                                             & 14.15                                                             & 8.99                                                             & 80.28                                                                  & 87.51                                                               \\
    NeRF-LOAM       & 2.93                                                             & \textbf{5.60}                                                     & 4.27                                                             & 92.07                                                                  & 94.22                                                               \\
    Ours            & \textbf{2.22}                                                    & \textbf{5.60}                                                     & \textbf{3.91}                                                    & \textbf{93.64}                                                         & \textbf{96.05}                                                      \\ \bottomrule
    \end{tabular}
    }
    \label{tab_maicity_quantitative}
    \vspace{-0.6cm}
\end{table}

In our first experiment, we evaluate the mapping quality on the simulated MaiCity dataset \cite{vizzo2021Poisson}, which provides both 64-beam noise-free LiDAR scans and a ground truth map of an urban scenario.
\Reftab{tab_maicity_quantitative} presents a quantitative comparison of our method and baselines. 
Our method achieves the best performance across all metrics.
An intuitive qualitative demonstration is shown in \Reffig{fig_qualitative}.
It is evident that our approach produces the most accurate and complete reconstruction.
For traditional methods, Voxblox \cite{oleynikova2017Voxblox} and Voxfield \cite{pan2022Voxfield}, the meshes obtained by them appear overly smoothed, lacking the preservation of details, like the tree and the pedestrian.
SHINE-Mapping \cite{zhong2023shine} uses a regularization-based continual learning strategy but still encounters the forgetting issue, resulting in an incomplete and non-smooth mesh.
NeRF-LOAM \cite{deng2023NeRFLOAM} achieves good ground reconstruction, due to its ground separation strategy.
Nonetheless, our method shows superior overall mapping accuracy, particularly in fine-grained structures and sparsely observed marginal regions.

\subsubsection{Newer College}
We also choose the real-world Newer College dataset \cite{ramezani2020newer} for evaluation.
This dataset comprises a hand-carried LiDAR sequence captured at Oxford University with noticeable measurement noise and motion distortion. 
A high-precision point cloud model observed by a terrestrial scanner is used as pseudo ground truth.
In accordance with NeRF-LOAM \cite{deng2023NeRFLOAM}, we take one out of every five scans for mapping in all methods.
This makes the task more challenging.
The quantitative results and qualitative results are shown in \Reftab{tab_ncd} and \Reffig{fig_qualitative}, respectively.
It can be seen that our approach outperforms all baseline methods on this noisy dataset.
Voxblox and Voxfield struggle with sparse observations and dynamic objects, producing large holes on the ground, and wrongly eliminating the tree region.
SHINE-Mapping and NeRF-LOAM can produce relatively complete maps.
But the former shows non-smooth reconstruction and large errors, even on flat ground.
The latter tends to produce fragmented artifacts in edge areas and small holes on the ground.
Our approach effectively balances reconstruction accuracy, completeness, and smoothness, yielding the best mapping performance.

\vspace{0.5cm}
\begin{table}[]
    \caption{Quantitative mapping results of different methods on the \emph{Newer College dataset}}
    \vspace{-0.2cm}
    \centering
    \setlength{\tabcolsep}{1.0mm}{
    \begin{tabular}{@{}ccccccc@{}}
    \toprule
    \textbf{Method} & \begin{tabular}[c]{@{}c@{}}\textbf{Acc.} $\downarrow$ \\ {[}$cm${]} \end{tabular} & \begin{tabular}[c]{@{}c@{}}\textbf{Comp.} $\downarrow$ \\ {[}$cm${]} \end{tabular} & \begin{tabular}[c]{@{}c@{}}\textbf{C-L1.} $\downarrow$ \\ {[}$cm${]} \end{tabular} & \begin{tabular}[c]{@{}c@{}}\textbf{Comp.Ratio} $\uparrow$ \\ {[}$20cm,$ $\%${]} \end{tabular} & \begin{tabular}[c]{@{}c@{}}\textbf{F-score} $\uparrow$ \\ {[}$20cm,$ $\%${]} \end{tabular} \\ \midrule
    Voxblox         & 8.73          & 12.17          & 10.45         & 89.85               & 90.81            \\
    Voxfield        & 8.32          & 10.82          & 9.57          & 91.17               & 91.27            \\
    SHINE           & 6.98          & 10.44          & 8.71          & 90.23               & 92.49            \\
    NeRF-LOAM       & 7.53          & 10.45          & 8.99          & 92.19               & 92.93            \\
    Ours            & \textbf{6.32} & \textbf{9.75}  & \textbf{8.04} & \textbf{92.86}      & \textbf{94.54}   \\ \bottomrule
    \end{tabular}
    }
    \label{tab_ncd}
    \vspace{-0.2cm}
\end{table}

\subsection{Ablation Study}

\begin{table}[tbp]
    \caption{The ablation study of our method on the \emph{Maicity dataset}}
    \vspace{-0.2cm}
    \label{tab_ablation}
    \setlength{\belowcaptionskip}{-4pt}
    \centering
    \begin{threeparttable}
    \setlength{\tabcolsep}{1.0mm}{
    \begin{tabular}{@{}ccc|cccc@{}}
    \toprule
    \textbf{Normal} & \textbf{Vox.} & \textbf{Hier.} &  \begin{tabular}[c]{@{}c@{}}\textbf{Acc.} $\downarrow$ \\ {[}$cm${]} \end{tabular} & \begin{tabular}[c]{@{}c@{}}\textbf{Comp.} $\downarrow$ \\ {[}$cm${]} \end{tabular} & \begin{tabular}[c]{@{}c@{}}\textbf{C-L1.} $\downarrow$ \\ {[}$cm${]} \end{tabular} & \begin{tabular}[c]{@{}c@{}}\textbf{F-score} $\uparrow$ \\ {[}$10cm,$ $\%${]} \end{tabular} \\ \midrule
    \ding{55}           & \ding{55}           & \ding{55}             & 3.03          & 8.28           & 5.66          & 91.87                                 \\
    \ding{51}           & \ding{55}           & \ding{55}             & 2.55          & 6.34           & 4.44          & 94.97                                 \\
    \ding{51}           & \ding{51}           & \ding{55}             & 2.31          & 5.91           & 4.11          & 95.82                                 \\
    \ding{55}           & \ding{51}           & \ding{51}             & 2.57          & 8.85           & 5.71          & 93.27                                 \\
    \ding{51}           & \ding{51}           & \ding{51}             & \textbf{2.22} & \textbf{5.60}  & \textbf{3.91} & \textbf{96.05}                        \\ \midrule
    Correction          & \ding{51}           & \ding{51}             & 2.40          & 7.00           & 4.70          & 94.42                                 \\
    \ding{51}           & KF                  & \ding{55}             & 2.81          & 6.45           & 4.63          & 94.78                                 \\ \bottomrule
    \end{tabular}
    }
    \begin{tablenotes}
        \footnotesize 
        \item ``Normal'' represents the option of normal-guided non-projective signed distance.
        ``Vox.'' represents the voxel-oriented sliding window.
        ``Hier.'' represents the hierarchical sampling strategy.
        ``Correction'' means applying distance correction with normals directly.
        ``KF''  means using the keyframe-oriented sliding window strategy.
    \end{tablenotes}
    \end{threeparttable}
    \vspace{-0.5cm}
\end{table}

We conduct the ablation experiments to verify the effectiveness of each component and provide both quantitative and qualitative results in \Reftab{tab_ablation} and \Reffig{fig_ablation}.
\subsubsection{Normal Guided Sampling}
We compare the performance of our method with and without normal guided sampling.
The results shown in \Reftab{tab_ablation} demonstrate that our normal guided sampling method significantly improves mapping accuracy and completeness.
The corresponding visual evidence can be seen in \Reffig{fig_ablation} (a1) and (b1).
The area highlighted by the orange box illustrates that our method manages to avoid overestimating the signed distance with large incidence angles, leading to more valid distance fields.
We further evaluate an alternative option that uses normals to correct the signed distance values for training \cite{deng2023NeRFLOAM}, as depicted in \Reffig{fig_ablation} (c1).
The results show that it can provide some improvement over using the projective distance, but our reconstruction is more continuous and complete.

\subsubsection{Voxel-oriented Sliding Window}
\begin{figure}
    \centering
    \includegraphics[width=0.99\linewidth]{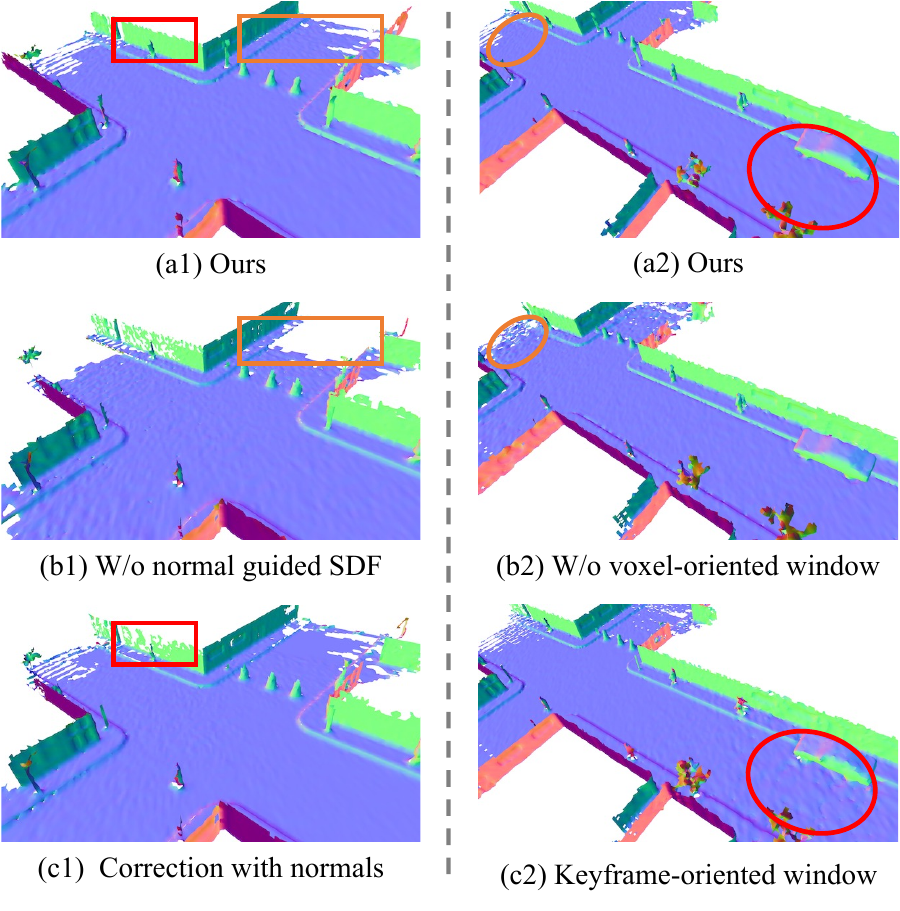}     
    \caption{Ablation study for our contributions and alternative designs on Maicity dataset.
    Regions are highlighted by colored boxes and circles to distinguish improvements}
    \label{fig_ablation}
    \vspace{-0.3cm}
\end{figure}

We adopt a naive replay-based method as the default baseline, which randomly samples data from all historical keyframes to train the implicit map together with current measurements.
\Reftab{tab_ablation} shows that the reconstruction errors significantly decrease when using the voxel-oriented sliding window.
This improvement is particularly evident in newly observed regions, as marked by the orange circle in \Reffig{fig_ablation} (a2) and (b2).
This is because our method can rapidly achieve convergence within the sliding window, while the replay-based baseline dilutes the training attention to newly observed regions by historical data.
We also provide the results of applying a keyframe-oriented sliding window for comparison.
As shown in \Reffig{fig_ablation} (c2), this design can enhance the reconstruction quality of the current local map.
However, it leads to inconsistent reconstruction in historically visited areas, as highlighted by the red circle, due to the forgetting issue.
In contrast, our method avoids this problem since each voxel retains complete supervision from various viewpoints.

\subsubsection{Hierarchical Sampling Strategy}
\begin{figure}
    \centering
    \includegraphics[width=0.99\linewidth]{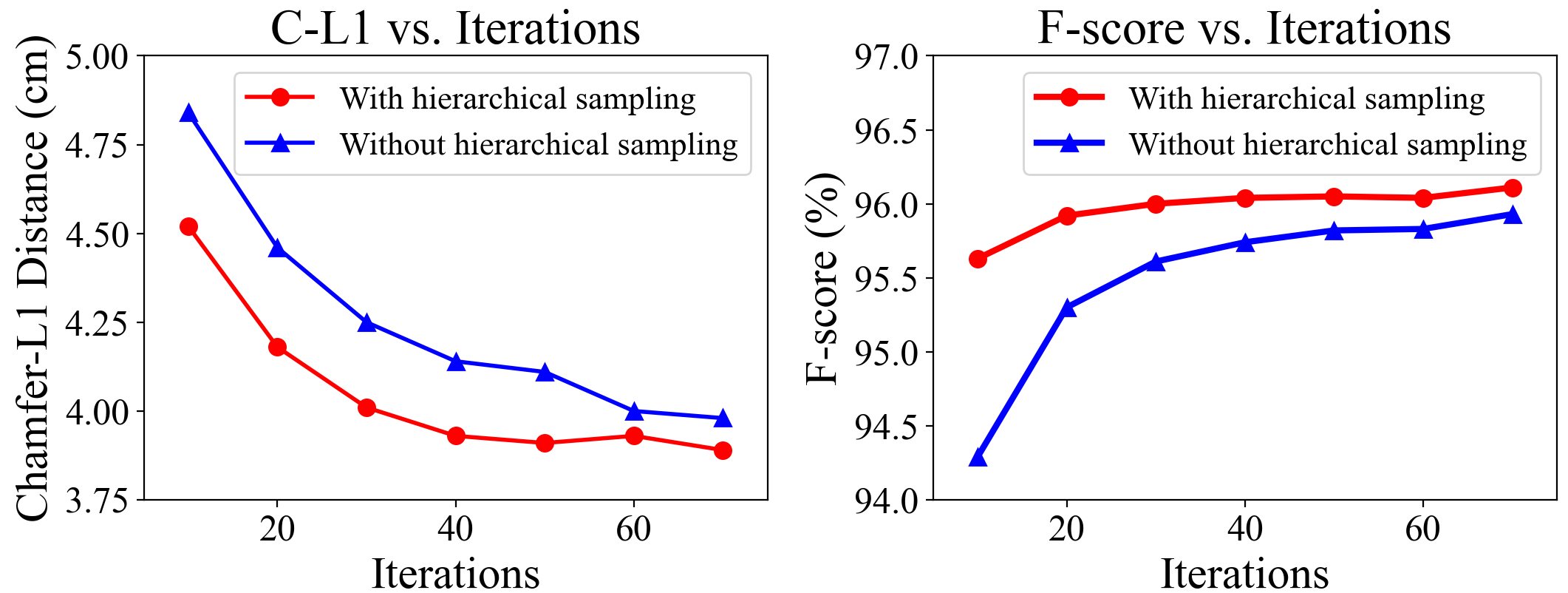}
    \caption{Ablation study of hierarchical sampling strategy.}
    \label{fig_ablation_hier}
    \vspace{-0.3cm}
\end{figure}

We compare the performance of our method with and without hierarchical sampling across different iteration settings to analyze its impact on training efficiency.
The results in \Reffig{fig_ablation_hier} show that as the number of iterations increases, mapping performance gradually improves, reaching convergence at around 40 iterations.
Notably, the mapping quality with hierarchical sampling is better even with fewer iterations.
This illustrates the effectiveness of this strategy in enhancing training efficiency.

\subsection{Scalable Incremental Mapping}

\begin{figure}
    \centering
    \includegraphics[width=0.99\linewidth]{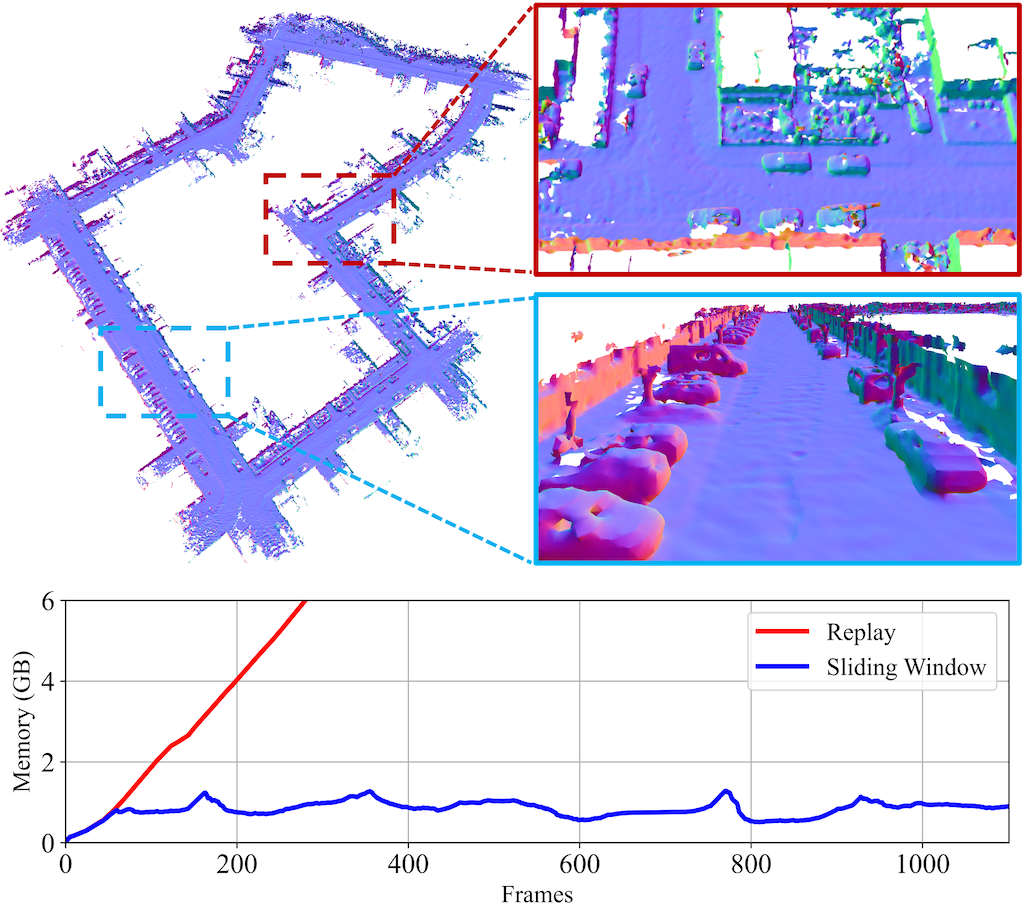}
    \caption{Top: the reconstruction result of KITTI odometry sequence 07. Bottom: the memory usage of historical data during the incremental mapping with and without our sliding window strategy.}
    \label{fig_incre_mapping}
    \vspace{-0.3cm}
\end{figure}

To demonstrate the scalability of our method to larger outdoor environments, we perform incremental mapping on the KITTI odometry dataset \cite{geiger2013kitti}.
\Reffig{fig_incre_mapping} showcases the reconstruction results of sequence 07 as an example.
It can be seen that our method produces a complete, consistent, and dense map within a real-world driving scenario.
Notably, thanks to our voxel-oriented sliding window strategy, the memory consumption of historical data remains stable at around 1GB throughout the mapping process.
In contrast, the replay-based strategy experiences rapid memory growth during mapping, eventually leading to memory overflow.

\subsection{Robustness Analysis}
\begin{figure}
    \centering
    \vspace{0.2cm}
    \includegraphics[width=0.95\linewidth]{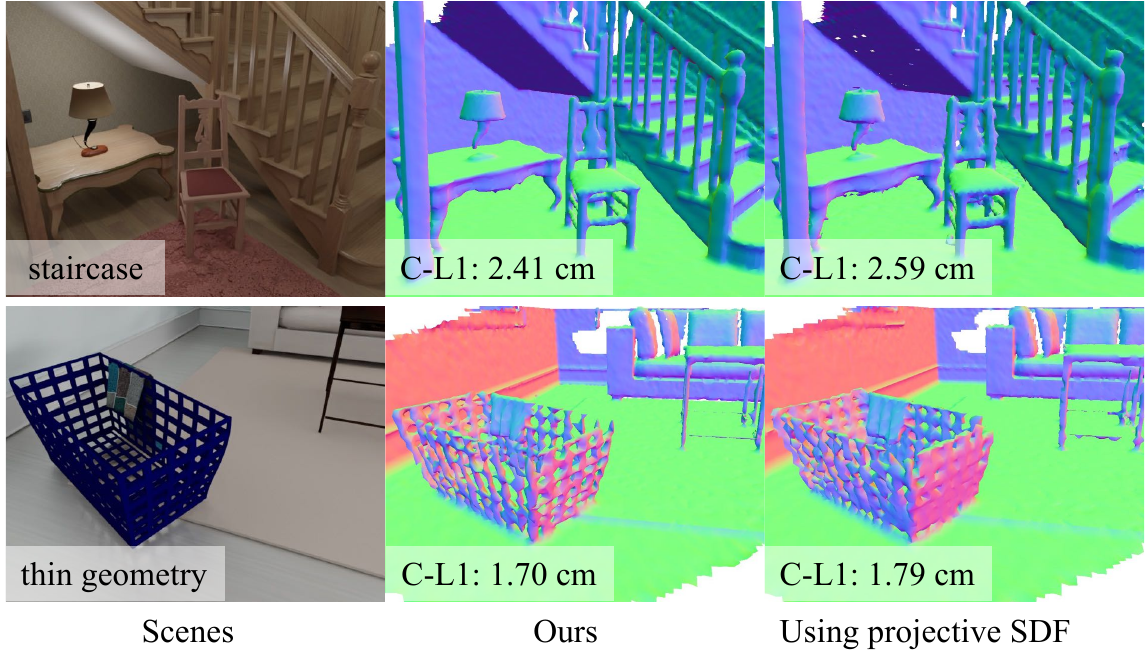}
    \caption{Indoor mapping examples of the \emph{Neural RGBD dataset}.
    Our approach manages to handle thin and complex geometries.}
    \label{fig_rgbd}
    \vspace{-0.3cm}
\end{figure}

\begin{figure}
    \centering
    \includegraphics[width=0.95\linewidth]{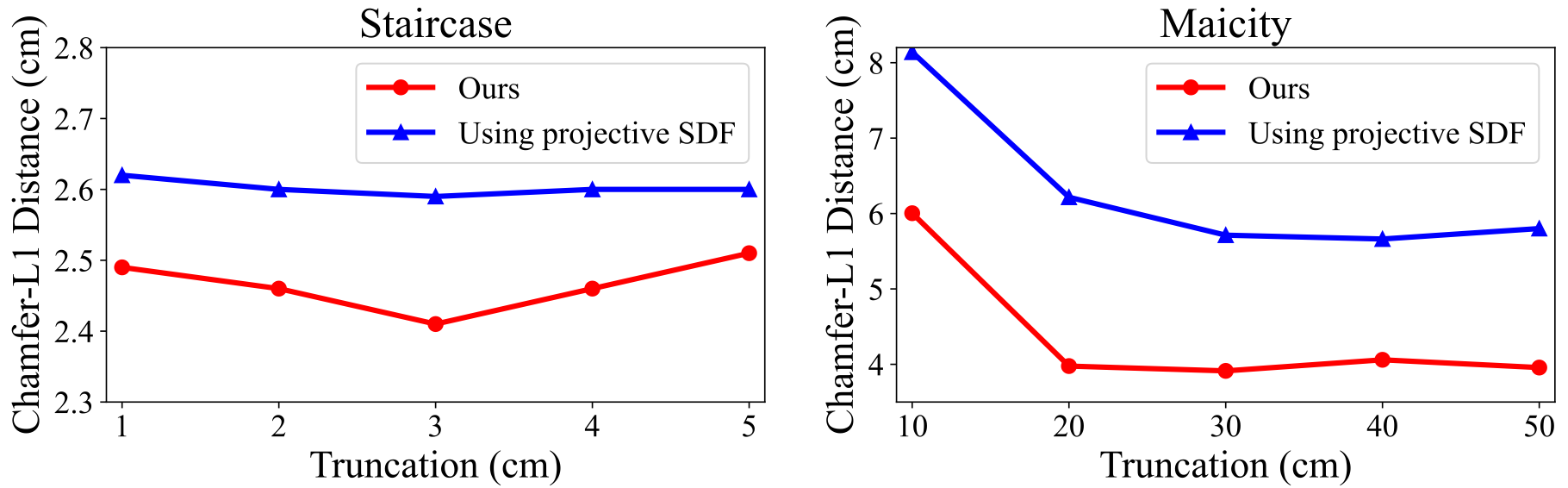}
    \caption{Impact of truncation distance on reconstruction quality.}
    \label{fig_ab_tr}
    \vspace{-0.5cm}
\end{figure}
We test our method on the Neural RGBD dataset \cite{azinovic2022Neural} to evaluate its robustness in different environments. 
Here we set the leaf voxel size to 4$cm$ and truncation distance to 3$cm$ to accommodate indoor environments.
\Reffig{fig_rgbd} demonstrates that our approach effectively captures intricate details of thin and complex structures, such as the chair and the crate.
As a reference, the results obtained by using projective distance exhibit a ``bulging effect" and suffer from noticeable artifacts.

\begin{figure}
    \centering
    \includegraphics[width=0.95\linewidth]{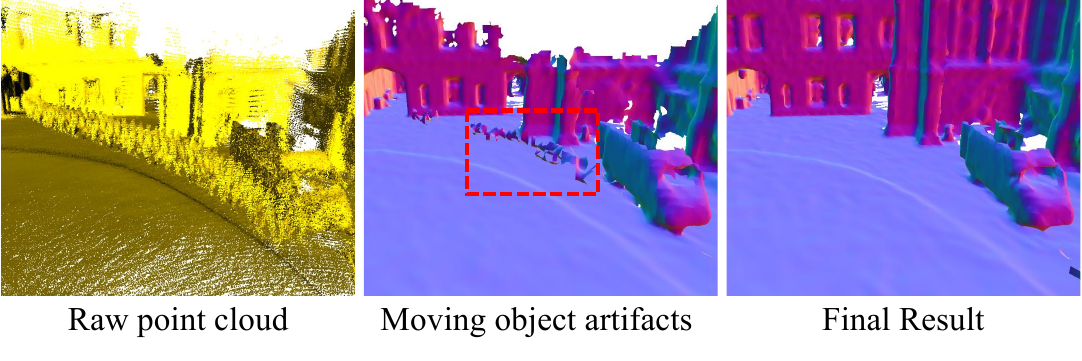}
    \caption{Impact of moving obstacles during the mapping process on the \emph{Newer College dataset}.
    Our approach manages to eliminate such dynamic objects.}
    \label{fig_dynamic}
    \vspace{-0.5cm}
\end{figure}

We also evaluate the impact of truncation distance on reconstruction quality.
As indicated in \Reffig{fig_ab_tr}, our method achieves lower C-L1 than those using projective SDF. 
However, our method also exhibits a slightly higher sensitivity to the truncation distance in the Staircase scenario with complex geometries.
This is because, in such cases, normal guided sampled points can easily interfere with other nearby surfaces, resulting in poorer reconstruction.
When applied in structurally simpler urban scenes (Maicity), our method demonstrates increased robustness to the choice of the truncation interval.
It is also worth noting that an excessively small truncation distance can compromise the completeness of the reconstruction, leading to a significant increase in overall error.

Lastly, we take a close look at the impact of moving obstacles on the mapping results, using the Newer College dataset as an example.
As depicted in \Reffig{fig_dynamic}, during the mapping process, moving objects like the pedestrians leave artifacts along their motion trajectories.
However, thanks to our free space sampling design, these artifacts can be gradually suppressed and eliminated, while static objects such as the van are retained.
Consequently, the final mapping results are complete and clean.

\subsection{Runtime Analysis}
We choose Maicity-01 and KITTI-07 to compare the runtime in \Reftab{tab_runtime}.
Although our method takes longer for training pairs allocation in the preprocessing stage, it avoids the time-consuming regularization update during training, resulting in better efficiency.
Currently, our method is unable to meet real-time requirements and performs incremental mapping offline.
It can be further sped up by the optimized C++ implementation and more efficient implicit representations \cite{liu2023RIM,muller2022Instant}.

\begin{table}[h]
    \caption{Average per-frame time consumption ($seconds$)}
    \vspace{-0.2cm}
    \centering
    \begin{tabular}{c|c|ccc}
    \hline
    \textbf{Dataset}         & \textbf{Method} & \textbf{Preprocess} & \textbf{Mapping} & \textbf{Total} \\ \hline
    \multirow{2}{*}{Maicity} & SHINE           & 0.13                & 2.37             & 2.50           \\
                                & Ours         & 0.25                & 1.84             & 2.09           \\ \hline
    \multirow{2}{*}{KITTI}   & SHINE           & 0.17                & 2.92             & 3.09           \\
                             & Ours            & 0.31                & 2.14             & 2.45           \\ \hline
    \end{tabular}
    \label{tab_runtime}
    \vspace{-0.3cm}
\end{table}

\section{CONCLUSION}
We introduce N$^{3}$-Mapping, a novel large-scale mapping approach using normal guided neural non-projective SDFs.
Experiments demonstrate that our method overcomes the limitations associated with projective distance and achieves more accurate and complete reconstruction.
Additionally, our voxel-oriented training strategy not only enables efficient incremental mapping without the forgetting issue but also ensures bounded memory consumption.
For future work, we can explore more robust non-projective SDF sampling methods.
Besides, apart from surface reconstruction, our implicit maps with their inherent continuity have the potential to be reused for accurate robot localization \cite{wiesmann2023LocNDF,deng2023NeRFLOAM}, which is the focus of our next work.

\bibliographystyle{IEEEtran}
\bibliography{reference}
\end{document}